\documentclass{article}

\usepackage[preprint, nonatbib]{neurips_2020}

\usepackage[utf8]{inputenc} 
\usepackage[T1]{fontenc}    
\usepackage{hyperref}       
\usepackage{url}            
\usepackage{booktabs}       
\usepackage{amsfonts}       
\usepackage{nicefrac}       
\usepackage{microtype}      
\usepackage{graphicx}
\usepackage{subcaption}
\usepackage{makecell}
\usepackage{amsmath}

\title{Unlocking the Potential of \\ Deep Counterfactual Value Networks}

\author{Ryan Zarick
  \\
  Minimal AI \\
  \And
  Bryan Pellegrino \\
  Minimal AI \\
  \And
  Noam Brown \\
  Facebook AI Research \\
  \And
  Caleb Banister \\
  Minimal AI \\
}
\newcommand{\plusMinus}{\raisebox{.2ex}{$\scriptstyle\pm$}}
\newcommand{\deepstack}{DeepStack}
\newcommand{\botname}{Supremus}

\DeclareMathOperator*{\argmax}{arg\,max}

\begin{document}

\maketitle
\begin{abstract}

Deep counterfactual value networks combined with continual resolving provide a way to conduct depth-limited search in imperfect-information games. However, since their introduction in the DeepStack poker AI, deep counterfactual value networks have not seen widespread adoption. In this paper we introduce several improvements to deep counterfactual value networks, as well as counterfactual regret minimization, and analyze the effects of each change. We combined these improvements to create the poker AI \botname{}. We show that while a reimplementation of DeepStack loses head-to-head against the strong benchmark agent Slumbot, \botname{} successfully beats Slumbot by an extremely large margin and also achieves a lower exploitability than DeepStack against a local best response. Together, these results show that with our key improvements, deep counterfactual value networks can achieve state-of-the-art performance.
\end{abstract}

\section{Introduction}
Imperfect-information games model strategic interactions between multiple agents that may have access to hidden information. Since an agent may not know what exact state they are in (due to the other agents' private information), many algorithms that are successful in perfect-information games, such as Q-learning and Monte Carlo tree search, are not sound in imperfect-information games. Instead, very different sets of algorithms are used.

The classic benchmark for zero-sum imperfect-information games is poker, and in particular no-limit Texas hold'em poker, which is the most popular form of poker played by humans. In 2017, both \emph{DeepStack}~\cite{moravvcik2017deepstack} and \emph{Libratus}~\cite{brown2017superhuman} claimed victory over human professionals in two-player no-limit Texas hold'em (HUNL). This was followed by \emph{Pluribus}~\cite{brown2019superhuman}, which defeated top humans in six-player no-limit Texas hold'em. These victories were considered major milestone achievements for the field of AI.

The key to victory for all these algorithms was lookahead search, but the way they conduct search is very different from what is used in perfect-information game algorithms.
A major challenge in search for imperfect-information games is determining what value to substitute at leaf nodes. In perfect-information games this is straightforward: the value at a leaf node is the estimated value of both players playing optimally from that point forward. However, since in imperfect-information games neither agent knows exactly what state they are in, the optimal policy for the remaining game is not well-defined when given only a state.

To address this challenge, \deepstack{} introduced \emph{deep counterfactual value networks} (CFVnets), which can in theory estimate the value of a state by conditioning the value of that state on the probability distribution over possible states that each player might be in~\cite{moravvcik2017deepstack}.
However, CFVnets did not become widely adopted.
State-of-the-art approaches instead search to the end of the game using an abstraction to reduce the size of the game beyond the depth limit~\cite{brown2017superhuman}, or pre-determine several policies for beyond the depth limit and solve the smaller meta-game where agents are limited to choosing one of those few policies for play beyond the depth limit~\cite{brown2018depth,brown2019superhuman}. This is true even for leading commercial poker assistance software~\cite{piosolver,simplepostflop}.

In this paper, we first present a reimplementation of \deepstack{} for HUNL and find that while it is not exploitable by a local best response~\cite{lisy2017eqilibrium}, it loses by a considerable margin to \emph{Slumbot}~\cite{slumbot}, a publicly available non-searching poker AI that was a top contender in the 2017 Annual Computer Poker Competition and the winner of the 2018 competition. We then introduce a series of improvements that, together, result in extremely strong performance. These improvements include an equilibrium-finding algorithm that achieves a far better empirical convergence rate than past algorithms, an efficient GPU-based implementation of this algorithm, more iterations resulting in much closer convergence to an equilibrium when conducting search, a larger search space, and far more training data. We measure the impact of each of these changes in the benchmark game of HUNL.
Finally, we combine these improvements into a HUNL bot called \emph{\botname{}} and show that it beats Slumbot, one of the strongest abstraction-based poker AIs ever developed, by a massive margin. We also show that it performs far better than \deepstack{} against a local best response.
\label{sec:intro}

\section{Notation and Background}
We consider a game with $N$ agents where the agents may have access to private observations. In this paper we generally assume $N=2$. However, empirically many of the algorithms perform well in some large-scale games involving more than two players~\cite{brown2019superhuman}.
The true state of the world, including the entire sequence of actions and all observations by any player, is denoted $s \in \mathcal{S}$ and is referred to as a \emph{world state} or \emph{history}. The set of world states that a player cannot distinguish between is referred to as an \emph{information state} (infostate) or simply \emph{state}. Finally, a \emph{public state} is a set of histories such that for any history $s$ in the public state, if $s$ and $s'$ share an infostate for any player then $s'$ is in the public state as well.

Typically, in imperfect-information games, the goal is to compute a \emph{Nash equilibrium} which is defined by a policy for each agent such that no agent can gain by deviating to a different policy~\cite{nash1950equilibrium}. Formally, a Nash equilibrium is a policy $\pi_i$ for each agent $i$ such that $v_i(\pi_i,\pi_{-i}) = \argmax_{\pi^*_i} v_i(\pi^*_i,\pi_{-i})$, where $\pi_{-i}$ is the policy of all other agents and $v_i(\pi_i,\pi_{-i})$ is the value to agent $i$ of playing policy $\pi_i$ against $\pi_{-i}$. In two-player zero-sum games, this is identical to a min-max equilibrium.

Computing an exact equilibrium in large-scale imperfect-information games is often intractable. Instead, one can approximate an equilibrium using an algorithm that ideally converges to an exact equilibrium. We measure the \emph{total exploitability} (or simply \emph{exploitability}) of policies $\pi_1$ and $\pi_2$ as $\argmax_{\pi^*_1} v_1(\pi^*_1,\pi_2) + \argmax_{\pi^*_2} v_2(\pi_1,\pi^*_2)$. In an exact equilibrium, exploitability is zero.

\subsection{Equilibrium-Finding Algorithms}
For small-scale imperfect-information games, linear programming methods can be used to precisely compute a Nash equilibrium. However, these techniques do not scale to extremely large games.

For large imperfect-information games, iterative algorithms that repeatedly traverse the game tree or portions of the game tree are used to approximate a Nash equilibrium. While there are many such iterative algorithms~\cite{brown1951iterative,leslie2006generalised,hoda2010smoothing,heinrich2015fictitious,kroer2018solving,kroer2018faster}, variants of the \emph{counterfactual regret minimization (CFR)} algorithm have shown the best empirical performance~\cite{zinkevich2008regret,tammelin2014solving,tammelin2015solving}. Variants of CFR have been used to solve heads-up limit Texas hold'em~\cite{bowling2015heads}, and to defeat humans in no-limit Texas hold'em~\cite{moravvcik2017deepstack,brown2017superhuman,brown2019superhuman}. The leading CFR variants are \emph{Discounted CFR (DCFR)} and \emph{Linear CFR (LCFR)}~\cite{brown2019solving}. DCFR is usually the better choice, though LCFR is better when there is a wide range of payoffs or when there is sampling involved.

For extremely large games, such as HUNL which has more than $10^{160}$ states~\cite{johanson2013measuring}, traversing the entire game tree with a CFR-like algorithm and updating the policy for each possible state is intractable.
Prior to 2017, the state-of-the-art approach to large-scale two-player zero-sum imperfect-information games (measured in the benchmark game of poker) was to first reduce the size of the game by bucketing together different states using techniques like k-means clustering and then approximating the solution for that simplified game. The solution for the game was stored as a lookup table for use during actual play against an opponent~\cite{johanson2012finding,johanson2013evaluating,jackson2013slumbot,bard2014asymmetric,brown2015hierarchical,brown2016baby,bowling2017heads}. Algorithms also exist that implicitly conduct abstraction by using function approximation instead of predetermining the abstraction~\cite{waugh2015solving,brown2019deep,steinberger2019single,li2020double}.

Around 2017, search algorithms that determine a policy for the next few moves in real time became popular~\cite{burch2014solving,jackson2014time,ganzfried2015endgame,moravcik2016refining,moravvcik2017deepstack,brown2017safe,brown2018depth}. These search algorithms were the key breakthrough that led to victories over humans in no-limit Texas hold'em poker.

\subsection{Counterfactual Value Networks}
Lookahead search has been a key component of superhuman AIs in perfect-information games such as backgammon~\cite{tesauro1994td}, chess~\cite{campbell2002deep}, and Go~\cite{silver2016mastering,silver2017mastering,silver2018general}. In these algorithms, whenever an agent must act it first generates a depth-limited \emph{subgame} consisting of the next several moves. At the end of the subgame are \emph{leaf nodes}, whose values are either known from the rules of the game (e.g., a checkmate in chess), or are determined via a value function. This value function takes as input the state of the game (e.g., the configuration of a chess board) and outputs an estimated value for each player assuming all players play optimally from that point forward. The value function may be a handcrafted heuristic or it may be learned from data.

In imperfect-information games, the input to a value function cannot simply be the state of the world, because optimal play also depends on the \emph{beliefs} of players about the state of the world, the other agent's beliefs about the state of the world, and so on. For example, the value to player~1 of holding the second-strongest poker hand when player~2 holds the third-strongest poker hand is very different depending on whether player~1 \emph{knows} that player~2 holds the weaker poker hand, or if player~1 thinks that player~2 holds a stronger poker hand. In the latter case, player~1 may simply fold and actually lose money.

One way to cope with this is to condition the value of a state on the beliefs of all players about the probability that each player is in each state. This concept was originally developed for partially observable Markov decision processes~\cite{nayyar2013decentralized}, and later independently developed for two-player zero-sum imperfect-information games and called counterfactual value networks (CFVnets)~\cite{moravvcik2017deepstack}. Since there are 1,326 combinations of private cards in HUNL, the input to the CFVnet would normally be 1,326 probabilities (summing to 1) for both players, the number of chips in the pot relative to the number of chips with which the players start, and the visible board cards. However, to reduce the complexity of the function approximation, the input size is reduced in both \deepstack{} and \botname{} by using bucketing.

Similarly, the output would normally be 1,326 values -- one for each hand the player could be holding. This too is reduced in both \deepstack{} and \botname{} by using bucketing.
One could alternatively add as an input the specific hand the player is holding, and then output only the value for that hand, but this is wasteful since determining the optimal policy in the subgame requires computing the policy for every feasible hand anyway.

\subsection{Continual Resolving}
\label{sec:resolving}
Another challenge of conducting lookahead search in imperfect-information games is that because the opponent's policy leading up to the current situation is unknown, the distribution over states at the start of the lookahead search is unknown. If any assumption is made that the opponent has a particular fixed policy, then the opponent could potentially adapt their policy to take advantage of that assumption.

A classic example of this is in rock-paper-scissors (RPS). Consider a sequential game of RPS in which player~1 acts and then player~2 acts without having observed the action of player~1. If player~2 conducts search and assumes that player~1 plays the equilibrium policy of choosing rock, paper, and scissors with $\frac{1}{3}$ probability each, then player~2 could reasonably decide to always choose rock. This has the same expected value as any other policy: zero. But if player~2 were to actually always throw rock, then player~1 could shift their policy to throwing paper more often.

\emph{Safe resolving}~\cite{burch2014solving,jackson2014time,moravcik2016refining,brown2017safe} copes with this challenge by adding constraints that the opponent's value for any state cannot improve in the new policy resulting from lookahead search compared to the value that was estimated from before lookahead search. Safe resolving provably bounds the exploitability of the lookahead search policy based on the accuracy of the value function.

Since the branching factor in HUNL is almost 20,000 actions, all competitive HUNL agents reduce the branching factor to a small number of raise sizes that are represented as fractions of the pot (typically in addition to fold, call, and raising all remaining chips). This is generally fine so long as enough actions are included in the action abstraction because a near-optimal policy can be computed that uses just a few actions in each situation. Indeed, professional poker players typically consider only one or two different raise sizes for a particular situation. However, if the opponent chooses a raise size that is not in the action abstraction then no policy is defined for responding to that action. Past poker bots coped with this through \emph{action translation}, in which the opponent raise size is rounded to a nearby in-abstraction size and treated as if it were that size~\cite{andersson2006pseudo,gilpin2008heads,schnizlein2009probabilistic,rubin2012case,ganzfried2013action}. However, action translation is potentially highly exploitable. Thus, bots that use action translation typically add a large number of action for the opponent in order to minimize the error due to action translation. This results in an exponential blowup in the size of the abstract game tree.

\emph{Continual resolving}, developed independently and simultaneously for both Libratus~\cite{brown2017safe} and \deepstack{}~\cite{moravvcik2017deepstack}, repeatedly applies safe resolving as play proceeds down the game tree. If an opponent chooses an action that is not in the lookahead search, continual resolving computes a policy in response to that action while attempting to ensure that the opponent's value for that action is no higher than any alternative action already in the lookahead search action abstraction. This eliminates any error due to action translation; the only potential for suboptimality comes from the opponent possibly choosing an action with a higher expected value than one that is already in the abstraction. However, this is in practice not a risk so long as a small number of good actions are added to the abstraction. The only downside is that continual resolving requires recomputing the policy potentially as often as every action rather than simply applying a lookup table.

\subsection{Description of Heads-Up No-Limit Texas Hold'em}
\label{sec:poker}
Heads-up no-limit Texas hold'em (HUNL) is a two-player zero-sum imperfect-information game. As is standard in the literature, we assume both players have \$20,000 in chips. The position of the players alternates after each hand. When it is a player's turn to act, they may either fold, call, or raise. If a player folds, they immediately lose the game and the other player receives all the money in the pot. If a player calls, they must put into the pot the same amount of money the other player has put in. If a player raises, they must first match the amount of money the other player has put in the pot, and then add any additional amount (so long as it is at least \$100, at least as big as any previous raise amount, and at most the player's remaining chips). A round ends when a player calls after another player has called or raised.

At the start of HUNL, both players are dealt two private cards from a standard 52-card deck. Player~1 must immediately place a \emph{small blind} of \$50 in the pot and player~2 must immediately place a \emph{big blind} of \$100 in the pot. A round of betting (called a \emph{round}) then occurs starting with player~1. The first round is referred to as the \emph{preflop}. After the preflop round ends, three community cards are dealt face up. At the end of the game, both players may use these community cards, in addition to their two private cards, to make their final hand. Next, another round occurs, referred to as the \emph{flop}, starting with player~$2$. Once the round is over, another community card is dealt face up and another betting round occurs called the \emph{turn}. Finally, one more card is dealt face up and the last round, called the \emph{river}, occurs. If no player has folded by the end of the river, then the player with the best five-card poker hand constructed from their two private cards and the five community cards wins the money in the pot. If the players tie then the pot is split evenly (i.e., both players receive a reward of $0$).

In the standardized form of HUNL used in the Annual Computer Poker Competition and past man-machine matches, the chip stacks for each player reset to \$20,000 after each game (hand). Performance is evaluated in milli big blinds per game (mbb/g), which is the average number of big blinds that an agent wins per 1,000 games.
\label{sec:background}

\section{Reimplementation of DeepStack}
\label{sec:validity}
Our reimplementation of \deepstack{} closely follows what was described in the paper. The neural networks were trained in the same manner using Adam~\cite{kingma2014adam} with Huber loss~\cite{huber1992robust} as an evaluation metric. For the reimplementation, all rounds except for the turn use a depth-limited lookahead tree identical to the one used in \deepstack{}.
In the \deepstack{} implementation, when a player is on the turn they solve until the end of the game using a bucketed abstraction for all river actions. The details of this bucketing were never presented, so instead we solve only to the start of the river and then use the river neural network from \botname{}. All architecture around the network inputs and outputs are identical to \deepstack{}.

The validation errors for the CFVnets in our reimplementation were 0.016 for the turn network (compared to 0.026 reported for \deepstack{}), 0.028 for the flop network (compared to 0.034 reported for \deepstack{}), and 0.000099 for the preflop auxiliary network (compared to 0.000055 reported for \deepstack{}). While the preflop auxiliary network has higher error, the errors are so small that the effect is likely minimal and greatly outweighed by the reduced errors on the flop and turn networks.

\begin{figure}[t]
    \centering
    \begin{minipage}{.48\textwidth}
        \begin{minipage}{1\textwidth}
          \centering
          \includegraphics[width=1\linewidth]{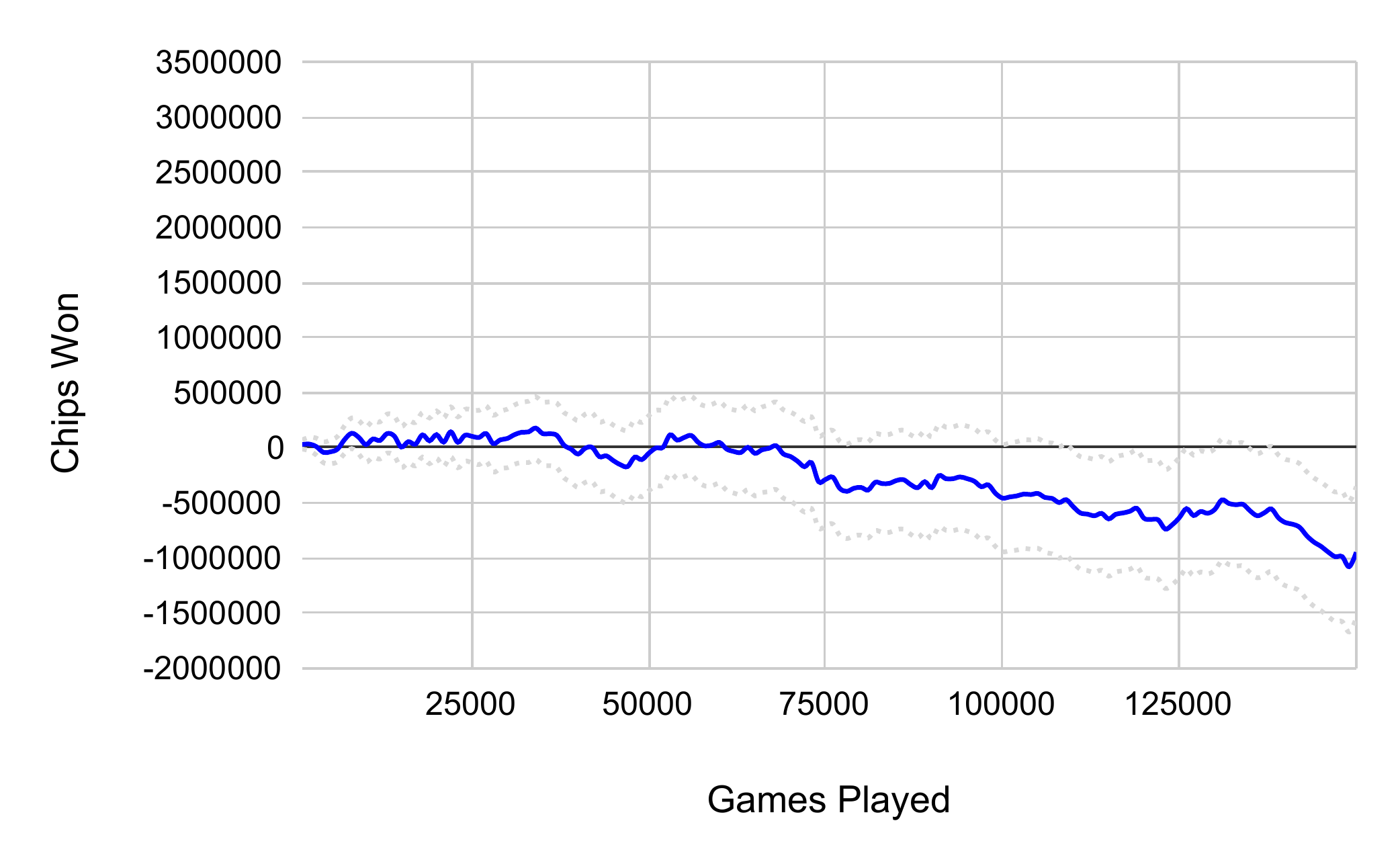}
          \captionof{figure}{Performance of \deepstack{} reimplementation versus Slumbot in chips won over 150,000 hands.}
          \label{fig:DeepstackVsSlumbotChips}
           \centering
              \includegraphics[width=1\linewidth]{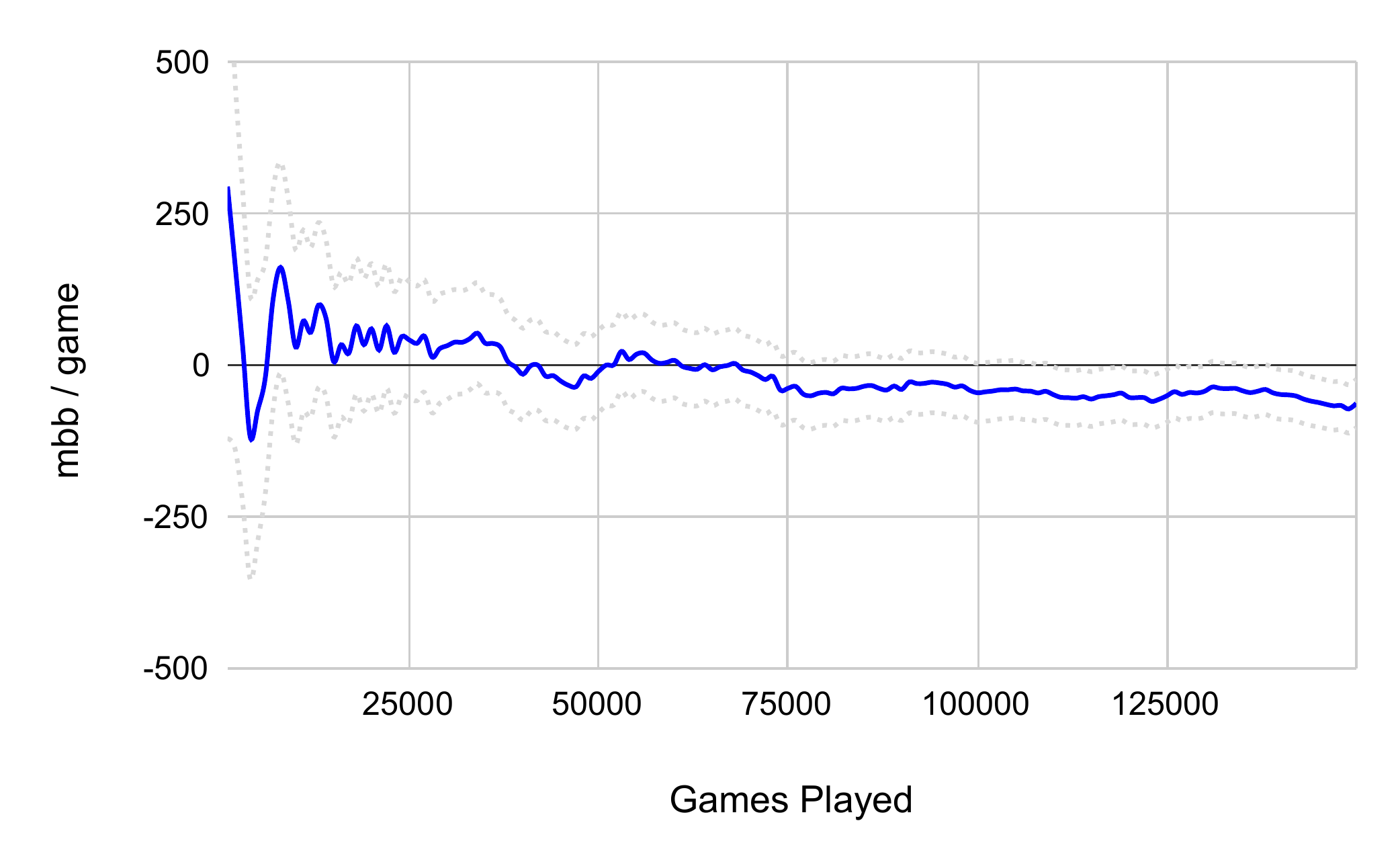}
              \captionof{figure}{Performance of \deepstack{} reimplementation versus Slumbot in mbb/g over 150,000 hands.}
              \label{fig:DeepstackVsSlumbotMbb}
              \includegraphics[width=1\linewidth]{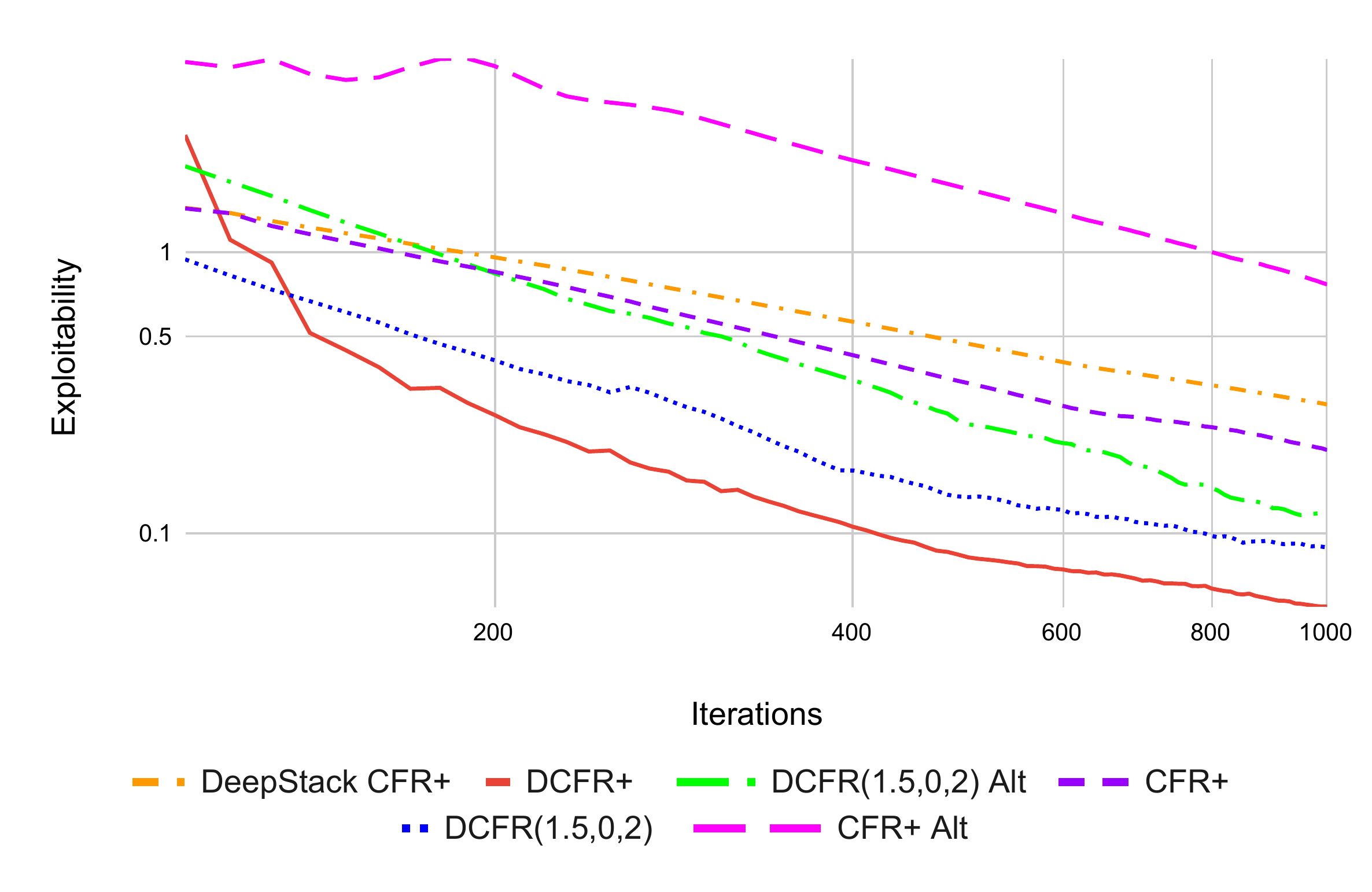}
                \caption{Performance of DCFR+ and popular CFR variants.}
                \label{fig:figure1}
            \end{minipage}%
         \begin{minipage}{1\textwidth}
             
        \end{minipage}
    \end{minipage}%
    ~
    ~
    ~
    ~
    ~
    \begin{minipage}{.5\textwidth}
      \centering
      \includegraphics[width=.75\linewidth]{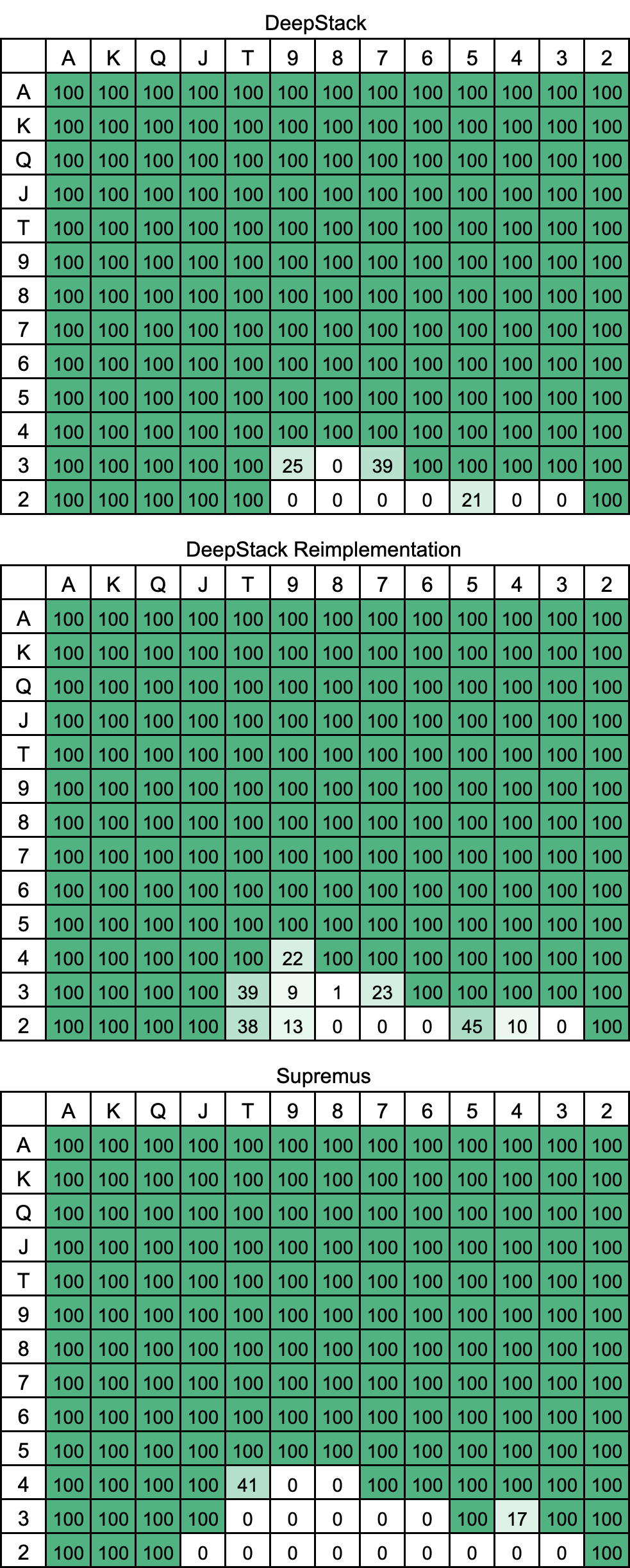}
      \captionof{figure}{Probabilities for not folding as the first action for each possible hand. The bottom-left half shows the policy when the suits of the two private cards do not match, and the top-right half shows the policy when the suits of the two private cards match.}
      \label{fig:frequencies}
    \end{minipage}
\end{figure}

To confirm that the reimplementation was similar to \deepstack{}, we looked at the frequencies of actions for various hands played by both \deepstack{} (in the 45,037-hand logs that were publicly released) and our reimplementation. Since many of the human opponents that \deepstack{} played against acted
far from what poker AIs and human professionals have well established as sound strategy, it was difficult to make accurate comparisons for situation in which an opponent had already acted. Therefore, we compared the action frequencies where the agent was the very first player to act and had no prior state influencing it. Figure~\ref{fig:frequencies} shows the frequencies are similar for both \deepstack{} and the \deepstack{} reimplemenation.

\deepstack{} was evaluated with two metrics: performance against human participants and performance against a local best response (LBR)~\cite{lisy2017eqilibrium}. While we cannot evaluate our reimplementation against the same human participants, we evaluate our reimplementation against LBR. Throughout this paper, we use \plusMinus{} to show the value of one standard error in results. In the \deepstack{} paper, when tested against LBR with fold and call as its only actions, \deepstack{} won 428 \plusMinus{} 87 mbb/g, while prior agents lost 400 mbb/g or more. When measured against LBR, our \deepstack{} reimplementation won 536 \plusMinus{} 68  mbb/g, which is slightly better than the original \deepstack{} result reports.

We also evaluated the \deepstack{} reimplementation in head-to-head performance against Slumbot, which is considered to be an extremely strong abstraction-based agent that was a top contender in the 2017 Annual Computer Poker Competition and winner of the 2018 competition. Slumbot does not conduct lookahead search; its entire policy is precomputed and used as a lookup table. This match used the standard HUNL format described in Section~\ref{sec:poker} that was used in both the ACPC and man-machine competitions. The two agents played each other for 150,000 hands. To reduce variance, all-in situations (in which both players have committed all their chips to the pot) that occurred before the final round were scored based on the expected value over all possible community card outcomes. This is a standard variance-reduction technique when evaluating win rates. The \deepstack{} reimplementation lost 948,096 chips to Slumbot, equating to a loss of 63 \plusMinus{} 40 mbb/g. The results can be seen in Figures~\ref{fig:DeepstackVsSlumbotChips} and~\ref{fig:DeepstackVsSlumbotMbb} respectively.

\section{Methods}

In this section, we introduce \botname{}, the  first HUNL agent using counterfactual value networks to demonstrate state-of-the-art performance against a leading agent. \botname{} is comprised of two major components: a sparse lookahead tree with continual resolving that is solved using an improved CFR variant, and neural network value functions used at the end of each round.

\botname{}'s neural network value functions were built with the same architecture as \deepstack{}, using a feedforward network with seven fully connected hidden layers and 500 nodes for each hidden layer, and an external network to enforce that the weighted averages of the outputs for each player sum to zero. The input to the network is a flattened array of length 2,001 which contains an encoded version of each player's probability distribution over private observations (poker hands) and public community cards compressed into a single array of length 1,000 and represented as probabilities over those buckets, as well as a final element containing the current pot size as a fraction of the starting number of chips. The network's outputs are the expected values of each of the 1,000 buckets, represented as a fraction of the current pot size.

Like \deepstack{}, \botname{} uses the CFR-D continual resolving algorithm described in Section~\ref{sec:resolving}. The lookahead tree uses a limited predefined set of actions, shown in Table~\ref{tab:sizes}, to build a game tree that extend to the end of the current round.
At the leaf nodes at the end of the current round, the value network is used to estimate the value at that point (except on the final round, which always extends to the end of the game).

Unlike \deepstack{}, \botname{} uses value functions at the end of the round for every round (except the final one). Without using a value function at the end of a round, lookahead trees require very coarse-grained action abstractions to be feasible to solve in real time because the number of nodes in a tree grows at a rate of $O(n^d)$ (where $n$ is the number of actions and $d$ is the depth of the tree) and the amount of time it takes to solve a tree scales linearly with the number of nodes in the tree. Additionally, at the end of each round in poker there is a large branching factor due to chance (due to additional community cards being revealed). Using a value function at the end of the round before any additional actions occur  eliminates the rapid growth due to chance and future actions.

As described in Section~\ref{sec:validity}, \deepstack{} used LBR to compare itself against other agents. In the \deepstack{} paper, when tested against LBR with fold and call as it's only actions, \deepstack{} won 428 \plusMinus{} 87  mbb/g, while prior agents lost 400 mbb/g or more. When measured against LBR, our \deepstack{} Reimplementation won 536 \plusMinus{} 68  mbb/g,  and \botname{} won 951 \plusMinus{} 96 mbb/g, almost doubling the winnings of \deepstack{} against the same configuration.

 \botname{} has several improvements over existing methods, including our modified CFR variant DCFR+, neural networks on all lookahead trees, and a complete end-to-end GPU implementation. For the remainder of this paper, we will describe each aspect in detail as well as benchmark them against existing approaches. 
  
\subsection{DCFR+}
In this section we introduce DCFR+, a modification of DCFR~\cite{brown2019solving} that achieves faster convergence over practical time horizons by delaying the starting point at which the average policy is tracked, as originally proposed in~\cite{tammelin2014solving}, and by using a linear weighting scheme for the average policy rather than quadratic. We also show empirically that when using deep counterfactual value networks, CFR converges faster when updating the regrets for both players simultaneously rather than alternating their updates.

Specifically, we modify the DCFR algorithm so that rather than having iteration $t$ receive a weight of $t^2$ when computing the average policy, it instead receives a weight of $\max\{0,t-d\}$ for some constant $d$. In our experiments, $d=100$. Since only a constant number of iterations receive a weight of zero and since the weights are non-decreasing, DCFR+ is still guaranteed to converge to a Nash equilibrium with a bound of $O(\frac{1}{\sqrt{T}})$ on exploitability, where $T$ is the number of iterations~\cite{brown2019solving}.

Figure~\ref{fig:figure1} measures the convergence rate of our DCFR+ algorithm against popular CFR variants, including the version used by \deepstack{}, when a counterfactual value network is used at the depth limit on the first betting round. Surprisingly, we observed faster convergence when updating both players simultaneously rather than alternating when a counterfactual value network is used at leaf nodes, while the opposite has historically been observed when not using counterfactual value networks. The best-known asymptotic convergence bound for both forms is similar~\cite{burch2019revisiting}.

\vspace{-0.1in}
\subsection{GPU}
\vspace{-0.1in}
\botname{} is a custom implementation written in CUDA with C++ to run entirely on a GPU. It has custom kernels optimized to achieve optimal performance and throughput.  It has a single on-ramp to load the initial tree into GPU memory and an off-ramp when DCFR+ is complete. This allows the software to never move memory in and out of the GPU, greatly reducing the memory latency during run-time. When comparing the speed of the two implementations using the same action abstraction on the flop, \botname{} completes 1,000 iterations in 0.8 seconds, over 6x faster than \deepstack{}. The faster GPU implementation combined with the use of DCFR+ results in \botname{} achieving an exploitability of 3 mbb/g over 5,000x faster on the same hardware.

\subsection{Neural Networks}
\botname{} uses neural networks for a value function at the start of each round (except the preflop, where an auxiliary value function is trained for the end of the round). The value network for each round was built by solving random subgames rooted at the start of that round, and using the value function of the round below for leaf nodes. Thus, the river network was trained first, followed by the turn network, followed by the flop network, and finally followed by the preflop auxiliary network. The random subgames were generated in a manner identical to \deepstack{}.
Each subgame was solved with 4,000 iterations per player of DCFR+.

The river network was trained on 50 million samples of randomly generated subgames, the turn network was trained on 20 million subgames, the flop network was trained on 5 million subgames, and the preflop auxiliary network was trained on 10 million situations. For comparison, \deepstack{} used 10 million examples for its turn network, 1 million for its flop network, and 10 million for its preflop auxiliary network (\deepstack{} did not use a river network).

As noted in Table~\ref{tab:huberloss}, the river network achieved an average Huber loss of 0.010 and 0.015 on the training and validation data, respectively. The turn network reached an average Huber loss of 0.008 on the training data and 0.010 on the validation data. The flop network achieved an average Huber loss of 0.0092 on the training and 0.011 on the validation data. It is important to note that the flop validation average Huber loss is three times lower than that of \deepstack{}'s loss. The auxiliary network was trained to 0.000069 and 0.000070 average Huber loss for the training and validation data, respectively. 

\begin{table} 
\caption{\botname{} versus \deepstack{} Huber loss results.}
\label{tab:huberloss}

\centering

\small

\begin{tabular}{l|rr|rr}
\toprule
 & \multicolumn{2}{c}{\botname{}}                                     & \multicolumn{2}{c}{\deepstack{}}                                 \\

\multicolumn{1}{c}{{\bf Network}}                         & \multicolumn{1}{c}{\bf{Trained}} & \multicolumn{1}{c}{\bf{Validation}} & \multicolumn{1}{c}{\bf{Trained}} & \multicolumn{1}{c}{\bf{Validation}}  \\
\toprule

\bf{River}                                        & 0.01                        & 0.015                          & \multicolumn{1}{c}{N/A}     & \multicolumn{1}{c}{N/A}         \\
\midrule
\bf{Turn}                                         & 0.008                       & 0.01                           & 0.016                       & 0.026                           \\
\midrule
\bf{Flop}                                         & 0.0092                      & 0.011                          & 0.008                       & 0.034                           \\
\midrule
\bf{Auxiliary}                                          & 0.000069                    & 0.000070                        & 0.000053                    & 0.000055    \\
\bottomrule                   
\end{tabular}
 \vspace{+0.08in}

\end{table}

\begin{table*}
\small
\caption{Actions used within each implementation's action abstraction. F, C, A refer to Fold, Call and All-In respectively while all remaining numbers refer to the fraction of the size of the pot being bet.}
\vspace{-0.1in}
\label{tab:sizes}
\begin{center}
\begin{tabular}{p{1.5cm}|p{3cm}p{3cm}p{1.75cm}p{1.5cm}}
\toprule
{\bf Name    }        & {\bf First Action }                                   & {\bf Second Action  }              & {\bf Third Action}  & {\bf Remaining } \\
\midrule
\deepstack{} & F, C, 0.5,1.0, 2.0, A                           & F, C, 0.5,1.0, 2.0, A        & F, C, 1.0, A  & F, C, 1.0, A      \\
\midrule
\botname{}      &\makecell{ F, C, 0.33, 0.5, 0.75,\\ 1.0, 1.25, 2.0, A} & F, C, 0.25, 0.5, 1.0, A      & F, C, 0.25, A & F, C, 1.0, A      \\
\bottomrule
\end{tabular}
\end{center}

\end{table*}

\subsection{\botname{} versus Slumbot}
\begin{figure}

    \centering
    \begin{minipage}{.48\textwidth}
      \centering
      \includegraphics[width=1\linewidth]{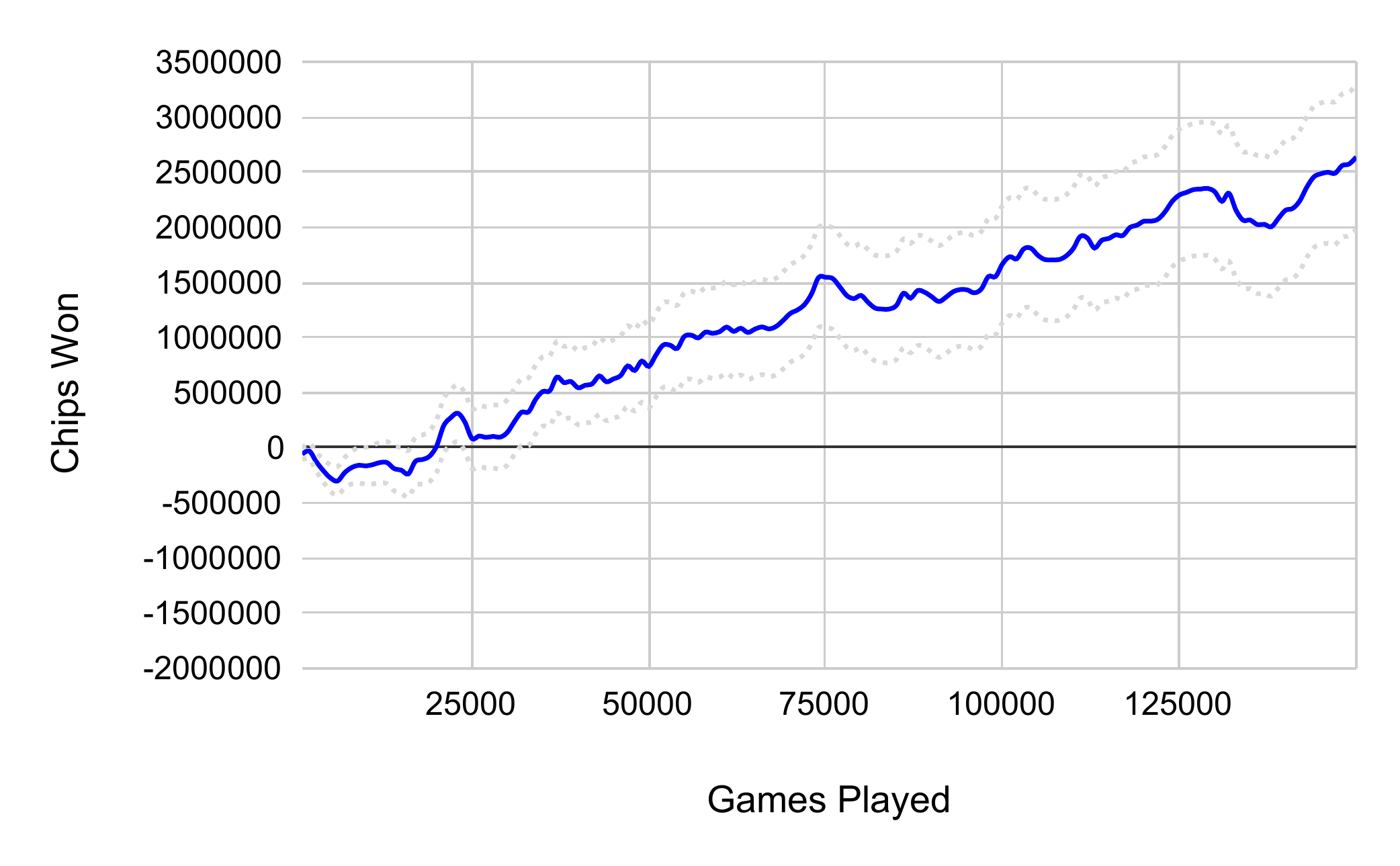}
      \captionof{figure}{Performance of \botname{} versus Slumbot in chips won over 150,000 hands. }
      \label{fig:UsVsSlumbotChips}
    \end{minipage}%
    ~
    ~
    ~
    \begin{minipage}{.48\textwidth}
      \centering
      \includegraphics[width=1\linewidth]{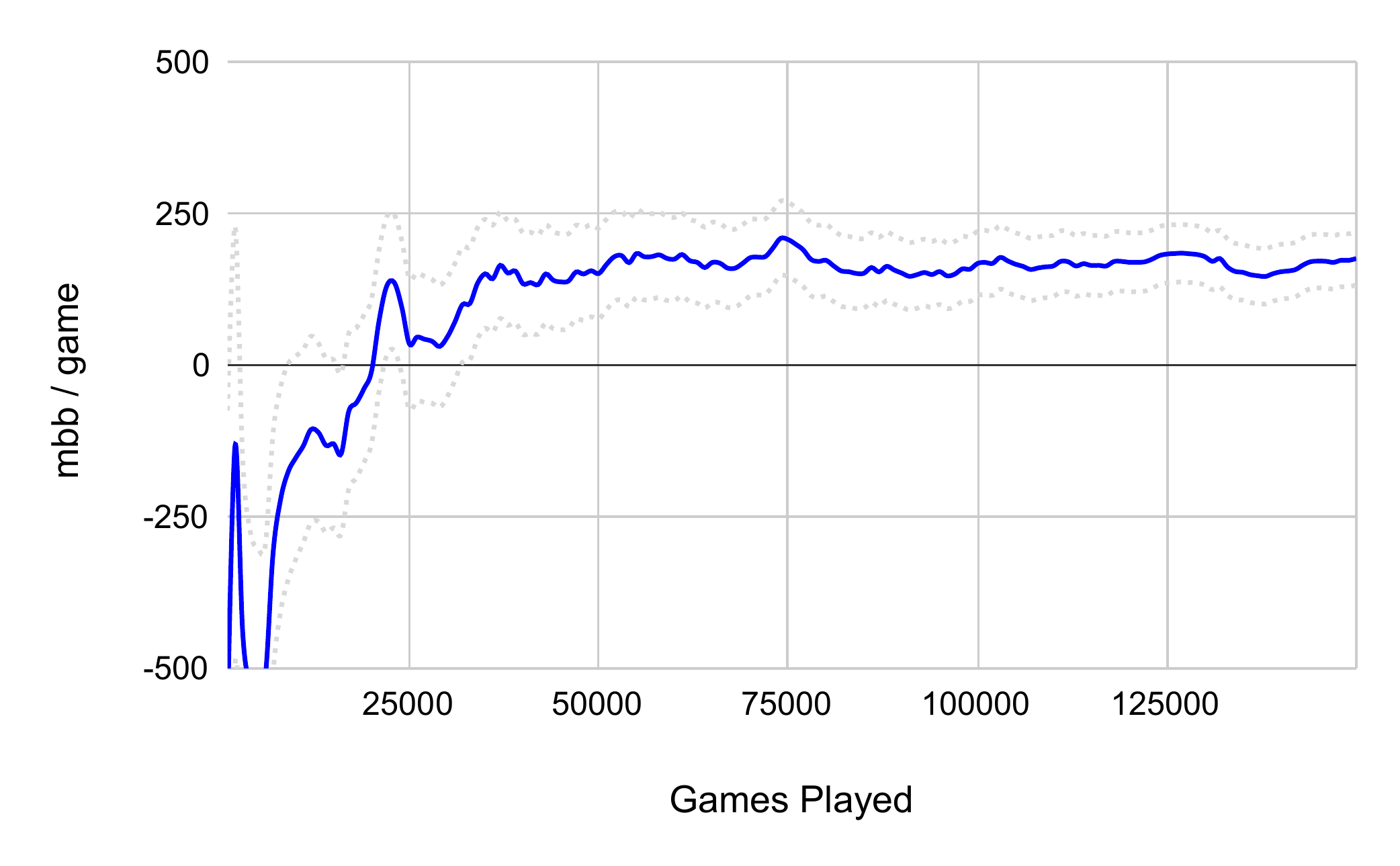}
      \captionof{figure}{Performance of \botname{} versus Slumbot in mbb/g over 150,000 hands.}
      \label{fig:UsVsSlumbotMbb}
    \end{minipage}
\end{figure}

\botname{} was evaluated by playing 150,000 hands against Slumbot, using the same time constraints as those used by \deepstack{}. This match used the standard HUNL format described in section~\ref{sec:poker}. \botname{} won 2,637,277 chips, equating to a win of 176 \plusMinus{} 44 mbb/g. These results can be seen in Figures~\ref{fig:UsVsSlumbotChips} and \ref{fig:UsVsSlumbotMbb} respectively.
Compared to the \deepstack{} reimplementation, which lost to Slumbot by 63 \plusMinus{} 40 mbb/g, our improvements resulted in an increase of 239 mbb/g against a strong benchmark agent.

\section{Conclusions}
Deep CFVnets were introduced more than three years ago, but have not seen widespread adoption. This may be due to the original deep CFVnet results not displaying superior head-to-head performance compared to alternative techniques. Indeed, we find that our reimplementation of \deepstack{} loses to the prior benchmark agent Slumbot by 63 \plusMinus{} 40 mbb/g. In this paper, we introduced a series of improvements to the construction and use of deep CFVnets, including more accurate CFR convergence, lower neural network error, and a finer-grained discretization of the action space. With these improvements, our agent \botname{} defeats Slumbot by 176 \plusMinus{} 44 mbb/g. These results show that achieving state-of-the-art performance with deep CFVnets is indeed possible with our key improvements.


\bibliography{main}

\bibliographystyle{plain}

\end{document}